\documentclass[journal]{IEEEtran}
\usepackage{cite}
\usepackage{support-caption}
\usepackage{subcaption}
\usepackage[utf8]{inputenc}
\usepackage{caption}
\usepackage{multirow}
\usepackage{blindtext}
\usepackage{tabularx}
\usepackage[table,xcdraw]{xcolor}
\usepackage{array}
\usepackage{url}
\usepackage{color}
\usepackage{booktabs}
\usepackage{times}
\usepackage{soul}
\usepackage{url}
\usepackage[hidelinks]{hyperref}
\usepackage{amsmath}
\usepackage{graphicx}
\usepackage{xspace}
\usepackage{xcolor}
\usepackage{fancyhdr}
\usepackage{graphics}
\usepackage{makecell}
\usepackage{csquotes}
\usepackage{xr}
\usepackage{tikz}
\usepackage{pgfplots}
\pgfplotsset{compat=1.16}
\usepackage{wrapfig}
\usepackage{subcaption}
\usepackage{enumitem}
\usepackage{ulem} % to strike through text
\usepackage{balance}
\usepackage[T1]{fontenc}% optional T1 font encoding
\definecolor{green}{rgb}{0.0, 0.5, 0.0}
\definecolor{amethyst}{rgb}{0.6, 0.4, 0.8}

\begin{document}
\title{Human Activity Recognition Using Cascaded Dual Attention CNN and Bi-Directional GRU Framework
%Recognizing Human Activity in Surveillance Videos using Encapsulated CNN and
%Bi-directional GRU Architecture
}

\author{Hayat~Ullah,~\IEEEmembership{Student Member,~IEEE,}
        Arslan~Munir,~\IEEEmembership{Senior Member,~IEEE,}
        % <-this % stops a space
\thanks{Manuscript received XYZ 29, 2022; Revised XYZ 22, 2022; Accepted XYZ; Published XYZ. This work was supported by the XYZ Project. (Corresponding author: Arslan Munir)
\newline
\indent H. Ullah and A.Munir are with the Intelligent Systems, Computer Architecture, Analytics, and Security Laboratory (ISCAAS Lab), Department of Computer Science, Kansas State University, Manhattan, Kansas, 66506, USA (e-mail: \href{mailto:}{hayatullah@ieee.org}, \href{mailto:}{amunir@ksu.edu}).
}
}
\maketitle

\begin{abstract}
Vision-based human activity recognition has emerged as one of the essential research areas in video analytics domain. Over the last decade, numerous advanced deep learning algorithms have been introduced to recognize complex human actions from video streams. These deep learning algorithms have shown impressive performance for the human activity recognition task. However, these newly introduced methods either exclusively focus on model performance or the effectiveness of these models in terms of computational efficiency and robustness, resulting in a biased tradeoff in their proposals to deal with challenging human activity recognition problem. To overcome the limitations of contemporary deep learning models for human activity recognition, this paper presents a computationally efficient yet generic spatial-temporal cascaded framework that exploits the deep discriminative spatial and temporal features for human activity recognition. For efficient representation of human actions, we have proposed an efficient dual attentional convolutional neural network (CNN) architecture that leverages a unified channel-spatial attention mechanism to extract human-centric salient features in video frames. The dual channel-spatial attention layers together with the convolutional layers learn to be more attentive in the spatial receptive fields having objects over the number of feature maps. The extracted discriminative salient features are then forwarded to stacked bi-directional gated recurrent unit (Bi-GRU) for long-term temporal modeling and recognition of human actions using both forward and backward pass gradient learning. Extensive experiments are conducted on three publicly available human action datasets, where the obtained results verify the effectiveness of our proposed framework over the state-of-the-art methods in terms of model accuracy and inference runtime across each dataset. Experimental results show that the proposed framework attains an improvement in execution time up to 167$\times$ in terms of frames per second as compared to most of the contemporary action recognition methods.
\end{abstract}
\begin{IEEEkeywords}
Convolutional neural network, channel-spatial attention, activity recognition, gated recurrent unit, pattern recognition, deep learning.
\end{IEEEkeywords}
\IEEEpeerreviewmaketitle
% Color brewer selection link
% https://colorbrewer2.org/#type=sequential&scheme=Blues&n=9
\section{Introduction}
\IEEEPARstart{T}{he} recent advancements in artificial intelligence (AI), in particular, deep learning-driven vision algorithms, and microelectronics have made possible automated surveillance on internet of things (IoT) and edge devices. Generally, these surveillance systems comprise of multiple interconnected cameras deployed in public places, such as big organizations, offices, roads, shopping malls, hospitals, and airports, to monitor humans and recognize their actions and behavior in video streams from multiple cameras. The primary objective behind the deployment of surveillance systems in the aforementioned places is to instantly detect abnormalities by recognizing the anomalous human behavior or activity in a video stream that can be harmful to the public. Human activity recognition is a process to analyze the hidden sequential pattern and predict the status of activity based on the perceptual context in input video stream. Generally, in videos, human activity is a combination of different movements of human body parts (i.e., hands, legs, or combination of both). For instance, running involves rapid movement of hands and legs, similarly, throwing object involves the backward and forward force of arm and hand. Human activity recognition has numerous potential applications, such as in smart surveillance systems \cite{huang2021abnormal}, video summarization \cite{sahu2021together}, content-based video retrieval \cite{qi2021semantics}, and human computer interaction \cite{ng2021multi}. In video, each frame contributes spatial information in sequential order which forms a sequential pattern containing human activity, that cannot be recognized in a single video frame. Considering throwing the knife and dart (that includes forward and backward force of arm and hand) have the same action pose in the starting frame; the discrimination of these two distinct activities becomes challenging while recognizing it with a single frame. Investigating the same movements of arm and hand in succeeding frames together with the information from previous frames will have the ability to effectively recognize human activities in video stream data. 

The earlier developed methods in initial research for vision-based activity recognition are exclusively focused on activities performed by a single person/actor in simple and controlled environment. However, the current research focuses on more challenging and realistic human activities recorded with clutter complex background, variation in viewpoint, occlusion in background, inter- and intra-class variations, and pose variations. More categorically, the existing vision-based human activity recognition methods can be categorized into two classes namely: (i) traditional handcrafted features-based, and (ii) deep learning-based human activity recognition methods. The traditional handcrafted features-based methods \cite{asghari2020online,ehatisham2019robust,naveed2019human,franco2020multimodal,elmadany2018information} use manually designed handcrafted or hand-engineered features (requires extensive human efforts with prior knowledge of scene understanding) followed by statistical machine learning models to recognize the activity. For instance, several traditional image features have been utilized to analyze videos, such as histogram of 3D oriented gradients (HOG3D), histogram optical flow (HOF) \cite{dileep2021anomalous}, motion boundary histograms (MBH) \cite{yenduri2022fine}, and extended speeded up robust features (SURF) feature descriptors. The hand-engineered features are required to be designed specifically for each particular environment based on scene perceptual complexity. Such type of manually designed handcrafted features-based methods fail to perform when using in an environment other than that for what these methods were designed. Recently, deep learning-based methods have made  incredible breakthroughs in various domain of image processing and computer vision, and have been actively used for human activity recognition problem \cite{luvizon2020multi,li2020spatio,ghose2020autofoley,lu2019gaim,liu2020multi}. These deep learning-based methods have obtained state-of-the-art performance by extracting deep progressive discriminative features using different convolutional neural network (CNN) kernels  and exploiting gradient learning strategy. Unlike, traditional handcrafted features of an image, deep CNNs learn progressively strong features (containing low-level, mid-level, and high-level features) that help to keep track of all type of visual semantics in image data.         

Deep learning-based methods have enhanced the activity recognition solutions in two perspectives. First, CNNs have the ability to extract more generic and semantically rich features than that of traditional handcrafted feature descriptors. Due to this generic feature extraction enabled by CNNs, CNNs have proliferated  in a variety of complex computer vision tasks including 3D image reconstruction \cite{hu20213dbodynet}, image and video captioning \cite{yan2021task}, and text-to-image generation \cite{xia2021tedigan} that cannot be accomplished using traditional handcrafted features-based methods due to their limitations in terms of features and learning strategies. Secondly, deep learning offers efficient architectures called recurrent neural networks (RNNs) which have the ability to learn the representation of human activity from a bunch of frames (sequence of frames or temporal representation of human activity) rather than a single frame. Earlier traditional methods consider frame-level classification of human activity in videos, rather than understating the activity in sequence of frames that greatly limits their performance for complex and multi-person activities. To cope with this challenge, deep learning-based methods have adopted RNNs for the better understanding and recognition of complex human activities in videos. Normally, in deep learning based-methods, RNNs are placed right after CNNs, where the CNN architecture is responsible for extracting deep discriminative features from videos and the RNN is responsible for learning the hidden sequential patterns in the extracted CNN features. The performance of these deep learning methods is good compared to traditional methods; however, these methods are computationally very expensive due to their hybrid and complex CNN and RNN architectures.  

The above-mentioned deep learning-based activity recognition methods have attained exceptional performance. Most of the existing AI-assisted activity recognition methods have adopted large yet effective pre-trained CNN architectures trained on a large-scale image dataset having tens of millions of trained parameters. Fusing such a computationally expensive feature descriptor backbone architecture with long short-term memory (LSTM) networks or multi-layer LSTMs (LSTMs having several layers with same settings) greatly increases the computational complexity of the overall method thereby compromising on the better tradeoff between model accuracy and complexity. Considering the demand for computationally-efficient yet effective approaches that provide a balanced tradeoff between model accuracy and complexity for deployment on resource-constrained IoT and/or edge devices, in this paper, we propose a deep learning based computationally efficient yet effective method for activity recognition problem that can be deployed even on resource-constrained edge devices in the IoT-enabled surveillance environment. Our main contributions in this work are as follows: 
\begin{enumerate}
    \item We propose a computationally efficient cascaded spatial-temporal learning approach for human activity recognition. The proposed system utilizes deep discriminative RGB features guided by channel-spatial attention mechanism and long-term modeling of action centric features for reliable recognition of human activities in video streams.
    \item We propose a light-weight CNN architecture having a total of 8 convolutional layers where the maximum number of kernels used per layer is 64 with spatial dimension of $3 \times 3$. With these constrained settings, we have developed a compact yet efficient CNN architecture for deep discriminative feature extraction as opposed to complex deep CNNs utilized by other contemporary works in their activity recognition models using transfer learning.
    %Analyzing action semantics at frame level is an essential task while recognizing complex human activities in video streams. For this purpose, numerous compute-intensive pretrained image classification CNN architectures have been utilized using transfer learning approach in the literature. Such cost-effective feature descriptor CNNs significantly increase the complexity of overall method and leading to the demands for high computational requirements. 
    \item We design a stacked dual channel-spatial attention mechanism with residual skip connection for spatial saliency extraction from video frames. The developed dual attentional module is placed after each two-consecutive convolutional layers of the developed CNN model which helps our network to extract saliency-aware deep discriminative features for localizing the action-specific regions in video frames.
    \item For efficient temporal modeling of long-term action sequences, we propose a bi-directional GRU network with three bi-directional layers (having forward and backward pass) that capture the temporal patterns of human actions in both forward and backward directions, which greatly enhances the reusability of features, improves the features propagation, and alleviates the issue of gradients vanishing. 
    \item We demonstrate the effectiveness and suitability of the proposed encapsulated dual attention CNN and bi-directional GRU framework (DA-CNN+Bi-GRU) for resource-constrained IoT and edge devices by comparing the model accuracy and execution/inference time of the proposed framework with various baseline methods as well as contemporary human action recognition methods.
\end{enumerate}

The remainder of this paper is organized as follows. Section ~\ref{sec:SectionRelatedWork} provides the brief overview on the related works covering different type of methods introduced for human activity recognition, till date. The proposed method and its technical component are discussed in detail in Section ~\ref{sec:proposedmethod}. In Section ~\ref{sec:experimentalresults}, we present extensive experimental evaluation of our method based on different assessment strategies. Finally, we conclude the paper in Section ~\ref{sec:conclusion} with possible future research directions.

\section{Related Works on Human Activity Recognition} \label{sec:SectionRelatedWork}
In recent years, human action and activity recognition have been widely studied and have got exceptional attention of computer vision researchers due to the recent success of deep learning for image classification and object detection task. Comprehensive review on both traditional and deep learning-based methods have been presented in numerous surveys \cite{pareek2021survey,kong2022human}. The reported literature on human action and activity recognition can be summarized in terms of handcrafted features-based methods, deep learning features-based methods, long-term temporal modeling-based methods, and attention models-based methods. This section presents a brief discussion on these representative methods and a brief summary of previous related works.

\textbf{\textit{Handcrafted features-based methods}} have been used to localize the spatial and temporal variations in videos using manually hand-engineered feature descriptors. Generally, these handcrafted features-based methods can be structured as a feature extraction and encoding pipeline having three phases including key features point detection (spatial and temporal feature points), quantization of detected features, and features encoding. The first phase involves the extraction of spatial-temporal features from video frames, followed by feature quantization in the second phase that quantize local motion-centric features. Lastly, the quantized spatial-temporal features are then encoded into feature vectors (known as action feature vectors) having fixed dimensions. For instance, inspired by the features extraction mechanism of the scale-invariant feature transform (SIFT) descriptor, Scovanner et al. \cite{scovanner20073} have adopted the SIFT algorithm features extraction strategy and extend their features space from 2D to 3D for encoding hidden action patterns. As single feature representation is not capable to capture human actions, therefore numerous multi-feature representative descriptors have been proposed in the literature. Laptev et al. \cite{laptev2008learning} have proposed a multiscale spatial-temporal features-based approach by utilizing space-time extension and Harris operator. They first extract multi-scale spatial-temporal features from video frames and then characterize appearance and motion of local features using volumetric histogram of oriented gradient. The retrieved multi-scale spatial-temporal features are then fed to non-linear support vector machine (SVM) for action recognition. In \cite{ryoo2016first}, Ryoo and Matthies have inspected the behavior of local and global motion features for recognizing first-person activities in video data. Their proposed methods exclusively focus on temporal structures depicted in first person action/activity videos. These traditional handcrafted features-based methods have shown progressive improvement over the years by presenting more efficient approaches; however, these methods are time consuming (lacking end-to-end recognition strategy), labor-intensive (requiring extensive human efforts to extract generic and more discriminative features), and difficult to adopt in diverse scenarios.

\textbf{\textit{Deep learning features-based methods}} are the current mainstream methods to solve the problem of complex human action and activity recognition in videos. With the recent success in computer vision domain for high-level vision tasks including image enhancement \cite{ullah2021light}, image segmentation \cite{chen2022saliency}, and video captioning \cite{aafaq2022dense}, CNNs have been actively investigated for human action and activity recognition problem. Where numerous CNN-assisted methods have been presented \cite{karpathy2014large,simonyan2014two,wang2015towards,yue2015beyond,wu2015modeling,wang2016actions,feichtenhofer2016convolutional}, having deep CNN architectures with 2D convolution kernels applied across convolutional layers of the CNN network. These convolutional layers extract deep discriminative spatial features with translation invariance from action video frames, offering reasonable action recognition performance without using temporal modeling. For instance, Karpathy et al. \cite{karpathy2014large} have presented a single-stage CNN architecture for action recognition, where they have trained their proposed model on a large-scale sport video datasets benchmark namely Sports-1M dataset. Although, their method acheives better results than traditional handcrafted features-based methods, the presented architecture is unable to cope with temporal modeling. To overcome this issue, several two-streams CNN architectures have been introduced \cite{simonyan2014two,wang2015towards,feichtenhofer2016convolutional} to obtain both spatial and temporal modeling of human action where one architecture performs spatial modeling of spatial contextual features and the second architecture performs temporal modeling using extracted optical flow features. The addition of second network improves the performance by introducing temporal modeling to CNN-based action recognition approach; however, it equally increases the computational complexity of the overall two-stream CNN approach. To achieve spatial modeling and temporal cues within a single CNN architecture without compromising on model complexity, 3D CNNs \cite{tran2015learning,varol2017long,ji20123d} have been introduced for human action recognition task. For instance, Tran et al. \cite{tran2015learning} exploits the powerful characteristics of 3D CNN to recognize human action in sports videos, where they have trained their proposed architecture on large-scale benchmark dataset and have shown promising results. However, these 3D CNN-based approaches work well with short-term temporal modeling and lack the ability to cope with long temporal modeling.

\textbf{\textit{Temporal modeling-based methods}} have been actively presented to  overcome the issue of long-term temporal modeling, where researchers have introduced a special kind of neural network called RNN, which has the ability to deal with the long-term sequences. Later, different variants of RNNs are introduced for action recognition problem including LSTM \cite{ullah2018activity}, bi-directional LSTM \cite{he2021db}, and GRU \cite{sun2022capsganet},  which are comparatively more efficient than RNNs in terms of memorizing contents for long period of time. For instance, Yue et al. \cite{yue2015beyond} have presented a two-stream CNN architecture to extract both spatial (edge, color and shape) features and temporal (optical flow) features stacked with LSTM model for temporal modeling of human activity. Similarly, Amin et al. \cite{ullah2018activity} have presented a two-stream CNN architecture followed by a multi-layer LSTM to recognize human activity in videos. They have first extracted spatial salient and optical flow features and then fed the extracted features to multi-layer LSTM for localizing human action in video sequences. Ibrahim et al. \cite{ibrahim2016hierarchical} have proposed a two-stream temporal modeling-based activity recognition framework to recognize a team or group of activities. Their proposed method consist of two LSTM networks, the first LSTM learn the representation of a single person action, whereas the second LSTM is responsible to understand collective activity by aggregating individual actions in a sequence of frames. Biswas et al. \cite{biswas2018structural} have presented a special variant of RNN named structural RNN for group activity recognition. Their proposed method consists of series of interconnected RNNs structured to analyze human actions and their mutual interactions in video sequences. To accurately learn the representation of human activity in feature-encoded video frames, Shugao et al. \cite{ullah2021efficient} have reformulated ranking loss to efficiently discriminate human activities. They have first extracted deep discriminative CNN features from video frames using VGG19, which are then fed into LSTM for analyzing hidden sequential patterns and recognition of human activities. Muhammad et al. \cite{muhammad2021ai} have presented a spatio-termporal approach for recognizing salient events in soccer videos, where they have used a pretrained ResNet50 architecture for deep features extraction and a multilayer LSTM for events recognition from the hidden sequential patterns. Although, these hybrid CNN+LSTM have shown significant performance for vision-based human action and activity recognition task, these methods are computationally complex due to intensive computation cause by CNN features extraction and human action modeling by LSTM.

\textbf{\textit{Attention mechanism-based methods}} have demonstrated great potentials for a variety of high-level vision tasks including image segmentation \cite{li2021abssnet}, video captioning \cite{deng2021syntax}, and visual questioning answering (VQA) \cite{yang2016stacked}. More recently, the attention mechanism combined with CNN and RNN networks have been widely used for human action recognition task and have achieved noticeable improvements in action recognition performance. For instance, Baradel et al. \cite{baradel2017human} have proposed a spatio-temporal attention-based approach for human action recognition, where they have exclusively focused on the tracking of human hands that helps to detect the discriminative segments of action in a video. They have used attention in recurrent style where they have embedded an attention mechanism in RNN network to efficiently model human actions. Islam et al. \cite{islam2021multi} have presented a multi-model graphical attention network for human action recognition which learn multi-model discriminative features. They have captured cross-modal relation using multi-model discriminative features extracted using message passing-based graphical attention mechanism. Long et al. \cite{long2018multimodal} have presented a method called keyless attention mechanism to effectively extract salient features, which are then fused with other extracted features to design multi-model features for human action recognition in videos. Song et al. \cite{song2018spatio} have presented a spatio-temporal attention model to examine spatial and temporal deep discriminative features for human action recognition in videos. They have used an LSTM network equipped with attention modules which is capable to exclusively focus on discriminative joints, and have applied multi-level attention on the joint-specific location. Moreover, they have also proposed a technique to model temporal action proposal for efficient action detection. Cho et al. \cite{cho2020self} have proposed a self-attention network for human action recognition which comprised of three different variants of self-attention network (SAN) named, SAN-V1, SAN-V2, and SAN-V3. Their developed models have the potential to extract high-level features by exploiting low-level correlation. Along with SAN model, they have also developed a temporal segment network (TSN) which greatly improved the overall action recognition performance of their proposed method. Although, these attention-driven methods have been widely used for human action recognition task and have obtained noticeable improvements over handcrafted features-based methods and other non-attention deep learning methods, these methods perform well only on clean red, green, and blue (RGB) video data and mostly fail while dealing with noisy color (RGB) video data.

\begin{figure*}[t]
    \centering
    \includegraphics[width=\textwidth]{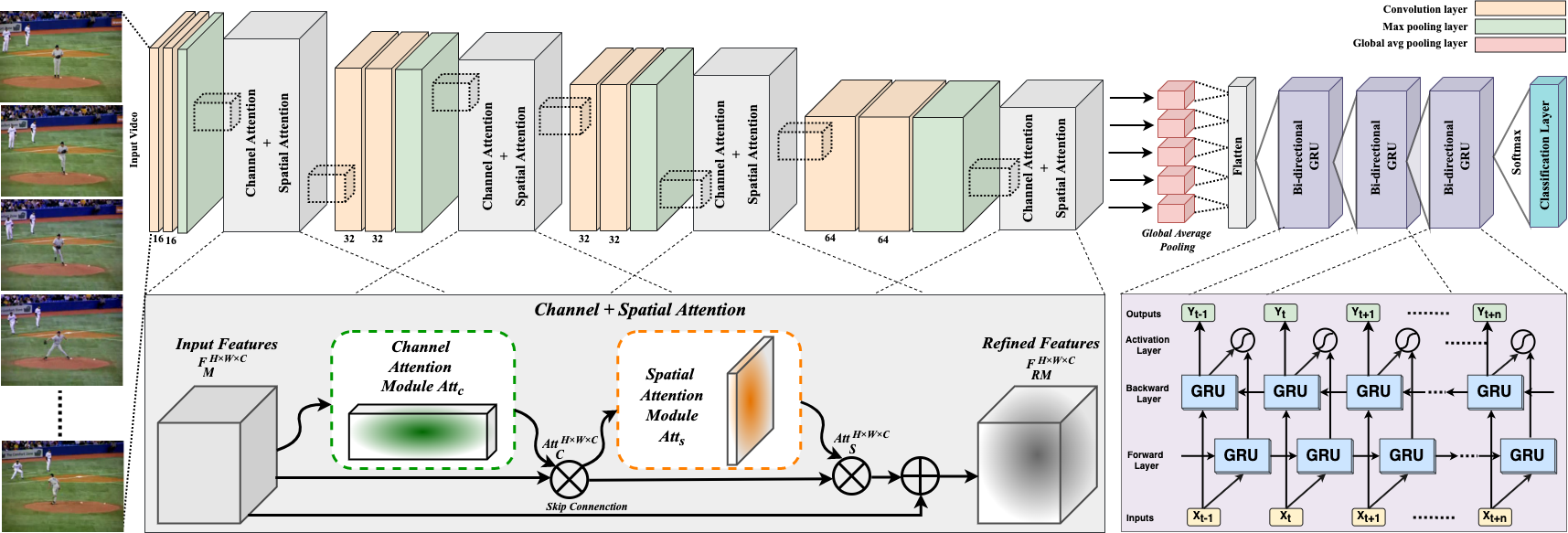}
    \caption{The graphical overview of our proposed activity recognition framework. The proposed framework consists of two main modules: an attention-driven CNN followed by a bi-directional GRU network. The CNN module utilizes a dual-attention mechanism to effectively extract salient CNN features from video frames, whereas the bi-directional GRU network is used to learn the activity representation for hidden sequential patterns.}
    \label{Fig:framework}
\end{figure*}

\section{Proposed Human Activity Recognition Framework} \label{sec:proposedmethod}
This section presents the detailed insights of our proposed human action recognition framework and its core components. For better understanding, the proposed approach is divided into three distinct modules, where each module is separately discussed. The first core component of our method is the newly introduced lightweight CNN architecture having a small number of trainable parameters. The second core component is a dual attention (channel and spatial attention) module, which is used to embed dual attention mechanism to the CNN module to enable our CNN model to extract salient features from video frames. The last key component of our framework is a bi-directional GRU network for learning long-term encoded patterns of human actions. The conceptual workflow of our proposed method is depicted in Figure 1.

\subsection{Overview of Proposed CNN Architecture}
Recognizing human actions in video data is indeed a challenging problem, where video data represent complex human actions over a series of frames in the form of different hidden visual contents that include temporal flow of objects in frames, varying texture, object-specific edges and colors. For better representation and modeling of human actions, these visual contents need to be analyzed effectively, which allows us to recognize the complex human actions or activity in video sequences. To effectively extract the defining visual features of these hidden action contents, CNN-based approaches are widely used to recognize human actions in videos. Although, the presented CNN-based approaches have shown remarkable performance, their computational complexity and execution/inference times are very high due to large network architectures. To avoid such high computational complexity and long runtime, we propose a tiny CNN architecture coupled with channel and spatial attention. The proposed CNN architecture contains a total of eight convolutional layers, where each two consecutive convolutional layers are followed by a max pooling layer and a dual attention block (containing both channel and spatial attention). The first two convolutional layers each apply 16 kernels on input video frames with the kernel size of $3 \times 3$, whereas the third and fourth convolutional layers each apply 32 kernels on the output of the first dual attention block with the kernel size of $3 \times 3$. Similarly, the fifth and sixth convolutional layers each apply 32 kernels on the output of the second dual attention block with the kernel size of $3 \times 3$. The last pair of the convolutional layers each apply 64 kernels on the output of the third dual attention block with the kernel size of $3 \times 3$ and then forward the estimated feature maps to the last dual attention block. The output of the last dual attention block is processed by a global average pooling layer, the output of which is then flattened by a flatten layer. The output of the flatten layer is fused with bi-directional GRU network for later long short-term sequence learning. The architectural details of our proposed CNN architecture are listed in Table \ref{tab:cnn}. It is worth noticing that we have used at most 64 convolutional kernels per layer and a fixed $3 \times 3$ kernel size that greatly help to reduce the computational complexity as low as possible with a negligible effect on model performance. 
\begin{table}[t]
\centering
\caption{Architectural details of our proposed CNN architecture.}
\resizebox{\columnwidth}{!}{
\begin{tabular}{c|c|c|c|c|c|c}\hline
Layer      &Input channels	&Number of kernels	&Kernel size  &Stride	& Padding & Output channels	 \\ \hline
Conv 1 	& 3 & 16 & $3 \times 3$ & 1 &1&16\\
Conv 2	& 16 & 16 & $3 \times3$ & 1 &1&16\\
\hline \multicolumn{7}{c}{\makecell{Max pooling}}\\
\hline \multicolumn{7}{c}{\makecell{\color{green}Channel Attention}}\\
\hline \multicolumn{7}{c}{\makecell{\color{orange}Spatial Attention}}\\ 
\hline
Conv 3	& 32 & 32 & $3 \times 3$ & 1 &1&32\\
Conv 4	& 32 & 32 & $3 \times 3$ & 1 &1&32\\
\hline \multicolumn{7}{c}{\makecell{Max pooling}}\\
\hline \multicolumn{7}{c}{\makecell{\color{green}Channel Attention}} \\
\hline \multicolumn{7}{c}{\makecell{\color{orange}Spatial Attention}}\\ 
\hline 
Conv 5	& 32 & 32 & 3$\times$3& 1 &1&32\\
Conv 6	& 32 & 32 & 3$\times$3& 1 &1&32\\
\hline \multicolumn{7}{c}{\makecell{Max pooling}}\\
\hline \multicolumn{7}{c}{\makecell{\color{green}Channel Attention}} \\
\hline \multicolumn{7}{c}{\makecell{\color{orange}Spatial Attention}}\\ 
\hline
Conv 7	& 32 & 64 & $3 \times 3$ & 1 &1&64\\
Conv 8	& 32 & 64 & $3 \times 3$ & 1 &1&64\\ 
\hline \multicolumn{7}{c}{\makecell{Max pooling}}\\
\hline \multicolumn{7}{c}{\makecell{\color{green}Channel Attention}} \\
\hline \multicolumn{7}{c}{\makecell{\color{orange}Spatial Attention}}\\ 
\hline \multicolumn{7}{c}{\makecell{\color{amethyst}Global Average Pooling}}\\ 
\hline \multicolumn{7}{c}{\makecell{Flatten}}\\ 
\hline
\end{tabular}}
\label{tab:cnn}
\end{table}
\subsection{Dual Attention Module}
To exclusively focus on most salient regions of video frames, we propose an attention-driven CNN architecture to efficiently localize the salient regions and enhance feature representation. The proposed attention mechanism is formed by fusing spatial attention module with the output of channel attention module through element-wise product operation. The detailed graphical overview of the proposed dual attention block is depicted in Figure 2. The fusion of both channel and spatial attentions not only helps to reduce the overall parameters overhead, but also enables the proposed CNN architecture to extract salient features. Therefore, we construct the formation of network layers in such way, where we place a stacked dual attention module after each two consecutive convolutional layers of our proposed network. The channel attention module estimates the weighted contribution of RGB channels by applying intermediate channel attention $\mathcal{A}_C$ on the output feature maps $F_{M}$ of the previous convolutional layer to obtain the channel attention $Att_C$. The computed output from $Att_{C}$ is then forwarded to the spatial attention module, which localizes promising object-specific regions by applying spatial attention on the computed channel attention feature maps $Att_{C}$. Finally, we obtain the refined feature maps $F_{RM}$ by fusing the spatial attention feature maps $Att_S$ with the input feature maps $F_M$ via a residual skip connection using element-wise addition operation. Mathematically, $Att_{C}$, $Att_{S}$, and $F_{RM}$  can be formulated as follows:  

\begin{equation} 
Att_{C}^{H\times W \times C} = \mathcal{A}_C(F_{M}^{H \times W \times C}) \otimes F_{M}^{H \times W \times C},
\label{eq:equation1} 
\end{equation}
\begin{equation} 
Att_{S}^{H \times W \times C} = \mathcal{A}_S(Att_{C}^{H \times W \times C}) \otimes Att_{C}^{H \times W \times C},
\label{eq:equation2} 
\end{equation}
\begin{equation} 
F_{RM}^{H\times W\times C}  = Att_{S}^{H \times W \times C} \oplus F_{M}^{H \times W \times C}
\label{eq:equation3} 
\end{equation}

\begin{figure}[t]
    \includegraphics[width=90mm ]{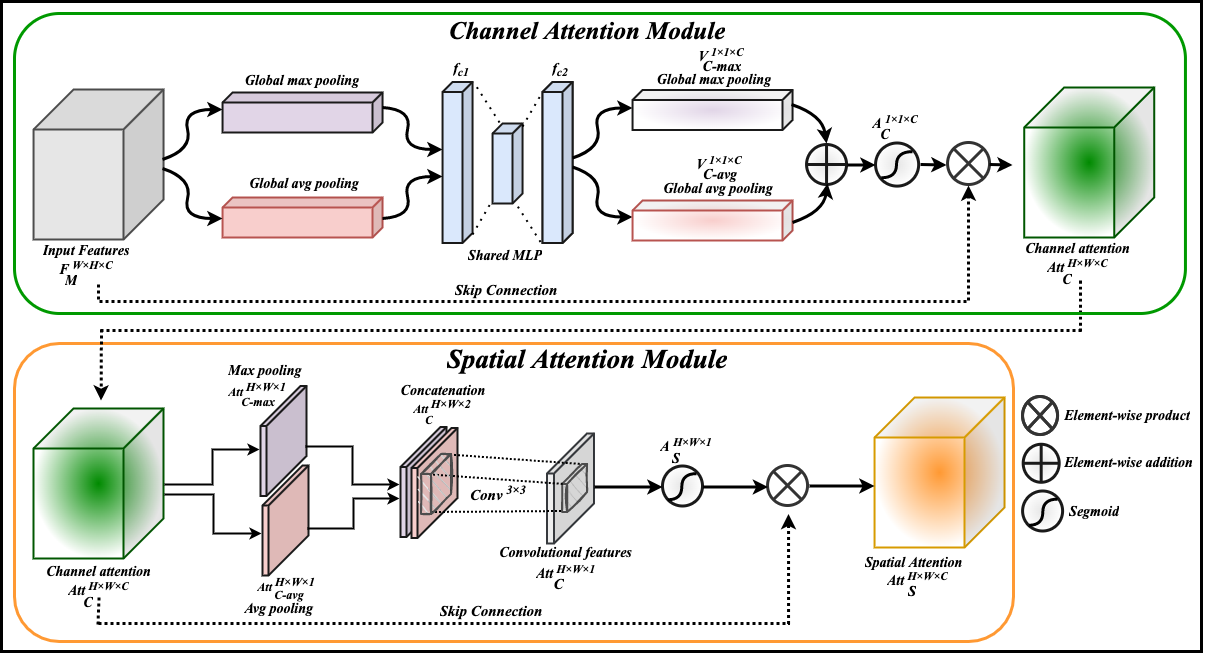}
    \caption{The detailed building block of dual attention block containing channel and spatial attention mechanisms.}
    \label{Fig:attention_block}
\end{figure}

Here, $\mathcal{A}_C$ and $\mathcal{A}_S$ are the intermediate channel attention and the intermediate spatial attention, respectively. $F_{RM}$ is the final refined feature maps obtained by fusing spatial attention and input feature maps $F_{M}$.
   
\subsubsection{Channel Attention}
In pattern recognition problems, particularly in image/object recognition, each color channel contributes differently based on the appearance of color in image. During training, a CNN model construct feature maps from input image data by extracting deep discriminative features over the number of convolution layers, where a particular channel contributes more than other channels in the recognition process. Unlike the earlier attention-based approaches that used either global max pooling layer or global average pooling layer, we have used both global max pooling and global average pooling to extract more effective features. The global max pooling emphasizes on highly activated values by selecting maximum value from the receptive field, where global average pooling estimates the equally weighted feature maps for each channel. 

The computed feature maps are then forwarded to a shared multilayer perceptron (MLP) containing two fully connected layers namely fc1 and fc2  having 128 and 512 nodes, respectively. The shared MLP learns the non-linearity between the two fully connected layers using ReLU activation function, and outputs two individual feature vectors namely $V_{C-max}^{1 \times 1 \times C}$  and $V_{C-avg}^{1 \times 1 \times C}$ for global max pooling and global average pooling, respectively. The computed feature vectors are then combined via an element-wise addition operation, and then forwarded to a sigmoid activation function, which normalizes the feature values to obtain intermediate channel attention features $\mathcal{A}_{C}^{1 \times 1 \times C}$. The obtained intermediate channel attention features $\mathcal{A}_{C}^{1 \times 1 \times C}$ are then fused with the input features maps $F_{M}^{H \times W \times C}$ using a residual skip connection by performing element-wise multiplication operation, which results in the ultimate channel attention feature maps $Att_{C}^{H \times W \times C}$ as depicted in Figure 2. Mathematically, the channel attention and its components can be expressed as follows:
\begin{equation} 
V_{C-max}^{1 \times 1 \times C} = fc2(R_{eLU}(fc1(maxpool(F_{M}^{H \times W \times C})))),
\label{eq:equation4} 
\end{equation}
\begin{equation} 
V_{C-avg}^{1 \times 1 \times C} = fc2(R_{eLU}(fc1(avgpool(F_{M}^{H \times W \times C})))),
\label{eq:equation5} 
\end{equation}
\begin{equation} 
\mathcal{A}_C^{1 \times 1 \times C} = \sigma(V_{C-max}^{1 \times 1 \times C} \oplus V_{C-avg}^{1 \times 1 \times C}),
\label{eq:equation6} 
\end{equation}
\begin{equation} 
Att_{C}^{H \times W \times C} = \mathcal{A}_C^{1 \times 1 \times C} \otimes F_{M}^{H \times W \times C},
\label{eq:equation7} 
\end{equation}

Here, $V_{C-max}^{1 \times 1 \times C}$ and $V_{C-avg}^{1 \times 1 \times C}$ are the obtained feature vectors from global max pooling and global average pooling operations, respectively. In the above equations, $F_{M}^{H \times W \times C}$ represents the input feature maps, $\sigma$ denotes the sigmoid activation function, whereas $Att_{C}^{H \times W \times C}$ is the final channel attention output.  

\subsubsection{Spatial Attention}
The spatial attention mechanism focuses on object saliency in the given feature maps by paying more attention to important features across each color channel and localizing salient regions. To highlight the salient object-specific regions in the feature maps, we exploit inter-spatial features and their relationship among channels, which greatly help to trace the target object in the feature maps. We compute the relation of inter-spatial features among channels by applying max pooling and average pooling to the input channel attention feature maps to obtain max-pooled channel attention $Att_{C-max}^{H \times W \times 1}$ and average-pooled channel attention $Att_{C-avg}^{H \times W \times 1}$, respectively.

The max-pooled channel attention $Att_{C-max}^{H \times W \times 1}$ and average-pooled channel attention $Att_{C-avg}^{H \times W \times 1}$ are concatenated and then forwarded to a single convolutional layer $Conv^{3 \times 3}$, which applies a $3 \times 3$ convolution kernel on pooled feature maps to form single-channel convoluted feature maps. These convoluted feature maps are then processed by a sigmoid activation function, which normalizes the learned features and produces intermediate spatial attention features $\mathcal{A}_{S}^{H \times W \times 1}$. Finally, the obtained intermediate spatial attention features $\mathcal{A}_{S}^{H \times W \times 1}$ are fused with the input channel attention feature maps $Att_{C}^{H \times W \times C}$ using a residual skip connection by preforming element-wise multiplication operation, which results in final spatial attention feature maps $Att_{S}^{H \times W \times C}$, as depicted in Figure 2. Mathematically, spatial attention $Att_{S}^{W \times H \times C}$ and its component can be expressed as follows:

\begin{equation} 
Att_{C-max}^{H \times W \times 1} = maxpool(Att_{C}^{H \times W \times C}),
\label{eq:equation8} 
\end{equation}
\begin{equation} 
Att_{C-avg}^{H \times W \times1} = avgpool(Att_{C}^{H \times W \times C}),
\label{eq:equation9} 
\end{equation}
\begin{equation} 
\mathcal{A}_S^{H \times W \times 1} = \sigma(Conv^{3 \times 3}(Att_{C-max}^{H \times W \times1} \biguplus Att_{C-avg}^{H \times W \times1})) ,
\label{eq:equation10} 
\end{equation}
\begin{equation} 
Att_{S}^{H \times W \times C} = \mathcal{A}_S^{H \times W \times 1} \otimes Att_{C}^{H \times W \times C} ,
\label{eq:equation11} 
\end{equation}
where $Att_{C-max}^{H \times W \times 1}$ and $Att_{C-avg}^{H \times W \times 1}$ are the global max and average pooled features, respectively. $\sigma$ is the sigmoid activation function and $\biguplus$ represents the concatenation operation. $Att_{S}^{H \times W \times C}$ is the final obtained spatial attention. The representative saliency maps of different human actions generated by our proposed method are depicted in Figure 3.     

\begin{figure}[t]
    \includegraphics[width=90mm]{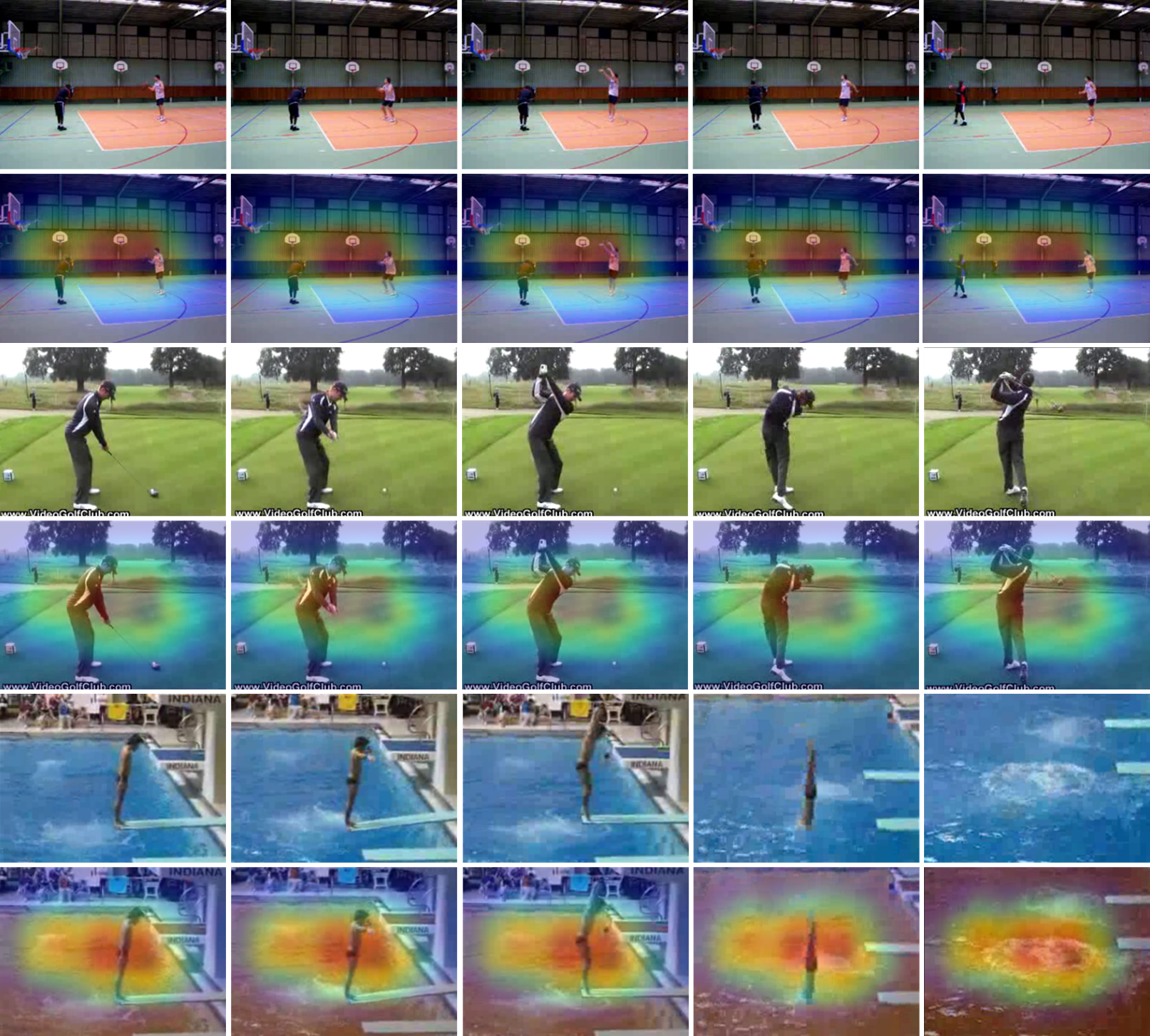}
    \caption{Visual representation of the computed salient object-specific regions by our dual  attention mechanism.}
    \label{Fig:vis_repre}
\end{figure}

\subsection{Learning Human Action Patterns via Bi-Directional GRU}
As videos are nothing but a stack of frames covering sequential flow of varying visual contents over a specific time interval. To understand the visual contents, main-stream computer vision approaches first extract deep discriminative features from the video frames using CNNs and then combine the extracted features in sequential order to maintain the semantic flow of the video. Second, the feature-encoded videos are then processed by RNNs to learn the representation of visual contents from hidden sequential patterns. Specifically, for human activity recognition problem, two special variants of RNNs are actively used by researchers that include LSTMs and GRUs. The LSTM unit comprises of different gates including input, output, forget gates and other memory components whereas the GRU unit contains an update gate, a reset gate, and an activation function. The LSTM is comparatively more complex than the GRU in terms of the number and formation of gates which leads to relatively higher computational complexity requiring more computational resources. Therefore, in this paper we propose to use GRU with bi-directional flow of learning strategy, which effectively learns from the encoded hidden sequential pattern.
The bi-directional GRU consists of two layers namely forward and backward layer, where both layers process the same sequence in different sequential order. The forward layer reads the input sequence from left to right, that is, from $X_{t-1}$ to $X_{t+n}$ where $n$ is the length of sequence. On the other hand, the backward layer reads the input sequence in reverse order from right to left, that is, from $X_{t+n}$ to $X_{t-1}$ as shown in Figure 4.  Both forward and backward GRU layers consist of GRU cells, where each cell consists of two gates namely a reset $r$ and an update gate $\mu$ with two activation functions that include sigmoid and tanh. The reset gate decides whether the GRU needs to forget or retain the portion of information based on its values (between 0 and 1). When the output value of reset gate is near to 0, the reset gate forgets the information from the previous portion of the sequence, whereas if the reset gate value is near to 1, the reset gate retains the previous portion of the sequence. The update gate decides the amount of information from the previous hidden state to be retained to the current hidden state based on its values (between 0 and 1). When the value of update gate is near to 0, the updated gate simply forgets the portion of information from the previous hidden state and retains the portion of information from the previous hidden state to current hidden state when the value is close to 1. Mathematically, the operation of these gates can be expressed as follows:

\begin{figure}[t]
    \includegraphics[width=90mm]{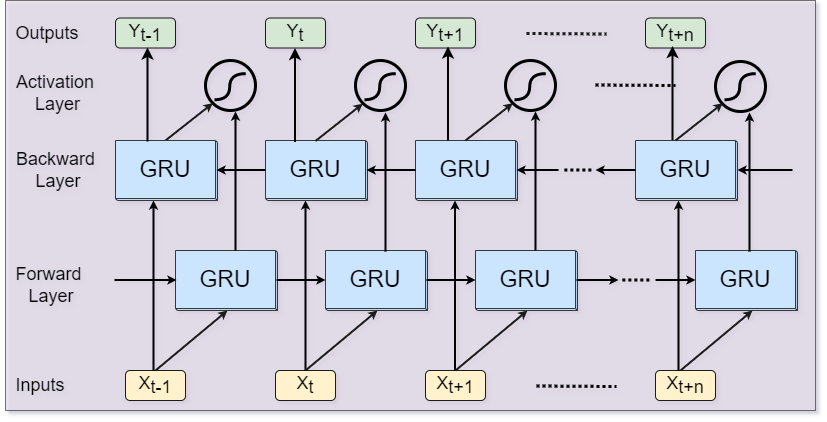}
    \caption{The building block of bi-directional single GRU layer.}
    \label{Fig:GRU_block}
\end{figure}

\begin{equation} 
r_{t} = \sigma(w_{r}\cdot x_{t} + u_{r}\cdot h_{t-1}),
\label{eq:equation12} 
\end{equation}
\begin{equation} 
\mu_{t} = \sigma(w_{\mu}\cdot x_{t} + u_{\mu}\cdot h_{t-1}),
\label{eq:equation13} 
\end{equation}
\begin{equation} 
\tilde{h}_{t}  = \tanh(w\cdot x_{t} + r_{t} \cdot u\cdot h_{t-1}),
\label{eq:equation14} 
\end{equation}
\begin{equation} 
{h}_{t}  = (1-\mu_{t}) \cdot h_{t-1} +  \mu_{t} \cdot \tilde{h}_{t},
\label{eq:equation15} 
\end{equation}
\begin{equation} 
{y}_{t}  = \sigma(w_{o} \cdot h_{t}),
\label{eq:equation16} 
\end{equation}

where $r_{t}$ and $\mu_{t}$ represent the reset and update gates, respectively, having values between 0 and 1. In the above equations, \emph{w} and \emph{u} are the weight variables, $x_{t}$ is the input to the GRU layer, $w_{o}$ is the weight variable between input and output layer, ${y}_{t}$ represents the output layer node at time step t. $\tilde{h}_{t}$ is the candidate hidden state of the current node, $\emph{h}_{t}$ is the current hidden state, and $h_{t-1}$ is the hidden state of the previous node. 
\color{black}

\section{Experimental Results and Discussion} \label{sec:experimentalresults}
In this section, we present detailed experimental evaluation of our proposed human activity recognition framework. We evaluate the effectiveness of our proposed framework by analyzing the performance with and without the key components (channel attention, spatial attention, bi-directional GRU) of our framework. First, we describe the implementation details and  performance evaluation metrics that we have used in this research. Next, we briefly discussed the datasets we have used for benchmarking experiments. We then compare our proposed framework with state-of-the-art human action recognition methods across each experimented dataset. Finally, we present the human action recognition visualization  and then conduct runtime analysis of our proposed approach for real-time human activity recognition. 

\subsection{Implementation Details}
The proposed framework is implemented using a well-known deep learning framework called TensorFlow version 2.0 in Python language 3 on a computing system with Intel Xeon (R) processor with processor frequency of 3.50\,GHz and having 32\,GB of dedicated main memory. The computing system is also equipped with an NVIDIA GeForce GTX 1080 graphics processing unit (GPU) having a graphics random-access memory of 8\,GB. For training and validation, we have divided the datasets into a ratio of 70\% and 30\%, where for training we have used 70\% of the data and the remaining 30\% of the data is used for validation. The training process is run for 300 epochs and the weights are initialized with a random weight initializer, whereas the batch size is set to 16. To adjust weight values during training, we have used the Adam optimizer with static learning rate of 0.0001. Our proposed network utilizes categorical cross entropy loss, which controls the weight adjustment based on network prediction during training. For sequence learning, we have used a sequence length of 16 frames  without overlapping for both forward and backward pass of bi-directional GRU, where we have used three bi-directional GRU layers with 32 GRU units per layer. Moreover, we have used two different performance evaluation metrics to assess the overall performance of our proposed method. The first metric is the \textit{accuracy} metric, which is used to evaluate the activity recognition performance of our framework and other contemporary methods. The second metric is \textit{frames per second (FPS)} or alternatively \textit{seconds per frame (SFP)}, which measures the runtime of our proposed framework and other contemporary methods. 

\subsection{Datasets}
To verify the effectiveness of our proposed framework, we have conducted extensive experiments on three challenging human actions datasets that include YouTube action, UCF50, and HMDB51 datasets. Each dataset consists of multiple action videos having varying duration, different view points, and frames per seconds (FPS). These datasets are discussed in detail in the following subsections.  

\subsubsection{YouTube Action Dataset}
The YouTube action dataset \cite{liu2009recognizing} is a commonly used action recognition dataset containing diverse sports and other action video clips collected from YouTube. The collected videos clips are very challenging due to variation in viewpoints, camera motion, cluttered background, and varying pose and appearance of objects in the scene. The dataset contains 1640 video clips categorized into 11 action categories, where the duration of videos range between 2 to 5 seconds having a frame rate of 29 FPS and a resolution of $320 \times 240$. The collected action clips in all action categories are grouped into 25 distinct groups containing 4 or more video clips, where each video clip in the same group share common visual features, such as background, viewpoint, and the person or actor.  
\subsubsection{UCF50 Dataset}
The UCF50 dataset \cite{reddy2013recognizing} is one of the challenging large-scale human activity recognition datasets, containing videos of diverse human actions captured with varying viewpoints, camera motions, object poses and appearances, and background clutter. The dataset contains a total of 6,676 video clips categorized into 50 different classes, where the duration of video clips range between 2 to 3 seconds with a frame rate of 25\;FPS and a resolution of $320 \times 240$. The video clips in all 50 categories are further grouped into 25 groups, where each group comprises of at least 4 video clips, where a video clip in a single group share common features of actions, such as the same person performing an action, the same view point, and the same background.  

\subsubsection{HMDB51 Dataset}
HMDB51 \cite{kuehne2011hmdb} is one of the challenging datasets commonly used for human action recognition in videos. The videos in this dataset are collected from difference sources including movies, public databases, YouTube, and Google videos. The dataset comprises a total of 6,849 action video clips categorized into 51 classes, where each class contains at least 101 video clips having duration of 2 to 3 seconds with a frame rate of 30\;FPS and a resolution of $320 \times 240$. The collected action video clips can be generally categorized into five different types of actions that include facial actions, facial actions with object manipulation, general body movements, body movements and interaction with objects, and body movements while interacting with humans.

\begin{figure*}[t]
\centering
\begin{subfigure}[t]{0.32\linewidth}
    \centering
    \includegraphics[width=\textwidth]{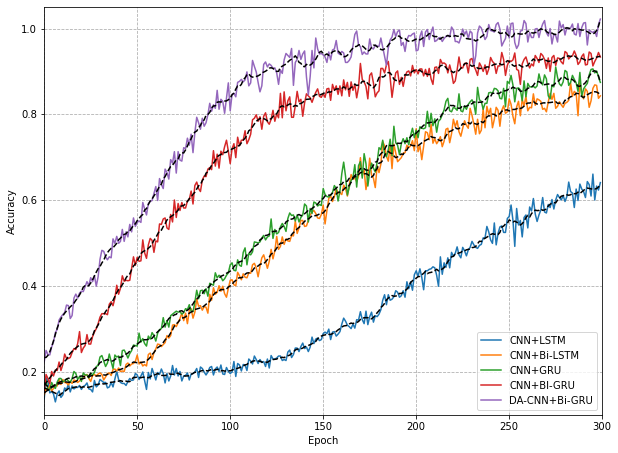}
    \caption{\centering}
\end{subfigure}
\begin{subfigure}[t]{0.32\linewidth}
    \centering
    \includegraphics[width=\textwidth]{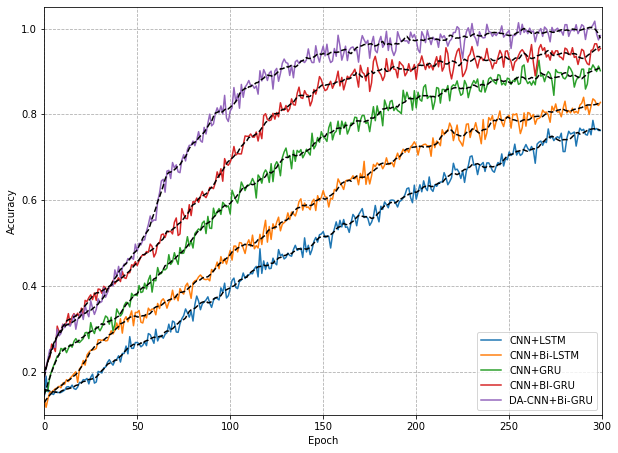}
    \caption{\centering}
\end{subfigure}
\begin{subfigure}[t]{0.32\linewidth}
    \centering
    \includegraphics[width=\textwidth]{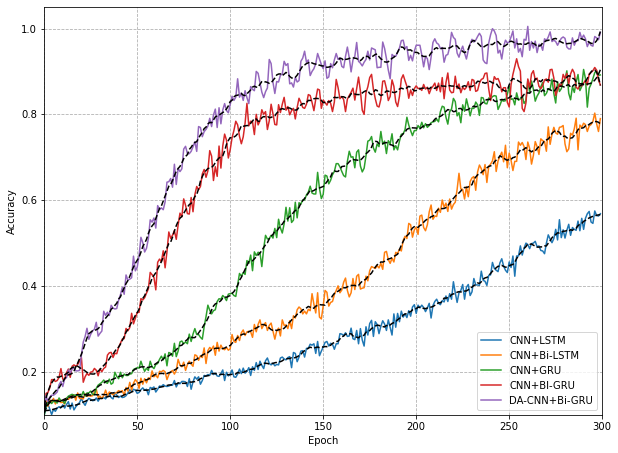}
    \caption{\centering}
\end{subfigure}
\caption{Training history of our proposed method along with other experimented baseline methods for 300 epochs over three benchmark action datasets: (a) Training history for YouTube action dataset, (b) Training history for UCF50 dataset, And (c) Training history for HMDB51 dataset.}
\label{Fig:TrainingHistory}
\end{figure*}
\subsection{Assessment of our Framework with Baseline Methods}
This research is built up on the exploration of various possible solutions for vision-based human action recognition, where we have developed several spatial-temporal methods, assessed their performances,  and developed our final proposed method. To obtain the optimal approach, we have explored different spatial-temporal solutions and successively developed four different baseline methods that include CNN+LSTM, CNN+Bi-LSTM, CNN+GRU, and CNN+Bi-GRU, and we have analyzed their performances in terms of model precision. To obtain a fair comparison, we have trained each baseline methods on three different datasets (i.e., YouTube action, UCF50, and HMDB51 datasets). These datasets are then used for training our proposed framework. The detailed network settings of these baseline methods are listed in Table \ref{tab:table2}, where it can be perceived that CNN+LSTM and CNN+GRU methods use a total of 11 spatial-temporal layers including 8 convolutional and 3 temporal layers. Similarly, CNN+Bi-LSTM and CNN+Bi-GRU methods use a total of 14 layers that include 8 convolutional and 6 temporal layers (having 3 forward and 3 backward pass layers). Finally, the proposed framework (DA-CNN+Bi-GRU) has a total of 18 layers comprising of 12 convolutional layers (8 convolutional and 4 attentional) and 6 temporal layers (having 3 forward and 3 backward pass layers). 

The training performance (in terms of accuracy) of each baseline method along with our proposed method is depicted in Figure~\ref{Fig:TrainingHistory}. It can be seen from Figure~\ref{Fig:TrainingHistory} that our proposed method (DA-CNN+Bi-GRU) performs better than other baseline methods in terms of accuracy. For instance, in Figure~\ref{Fig:TrainingHistory}\,(a) for YouTube action dataset, our method achieves the best accuracy score throughout 300 epochs. In Figure~\ref{Fig:TrainingHistory}\,(b) for UCF50 dataset, our method (DA-CNN+Bi-GRU) does not perform the best in early 35 epochs, where CNN+Bi-GRU dominates; however, after 35 epochs our methods starts improving and finally trains with the best accuracy at 300th epoch. Similarly, in Figure~\ref{Fig:TrainingHistory}\,(c), our method (DA-CNN+Bi-GRU) starts as the second-best method in early training epochs where CNN+Bi-GRU dominates; however, after 20 epochs our proposed method attains the best accuracy as compared to the other baseline methods and remains the best till the end of training. The obtained performances of these baseline methods along with our proposed method across three benchmark datasets are presented in Table \ref{tab:table3}. From Table~\ref{tab:table3}, it can be noticed that the proposed framework dominates all the baseline method across each dataset. For instance, the proposed framework attains the best accuracy score of 98.0\% over YouTube action dataset among all the baseline methods, whereas CNN+Bi-GRU obtains the second-best accuracy score of 92.1\%. Similarly, on UCF50 dataset, the proposed framework obtains the highest accuracy score of 97.5\%, whereas the runner-up is CNN+Bi-GRU with an accuracy of 93.6\%. Finally, for HMDB51 dataset, it can be seen that our proposed method dominates all the baseline methods by achieving the best accuracy score of 79.3\%, whereas CNN+Bi-GRU is the runner-up method that attains the second-best accuracy score of 72.4\%. The best and the runner-up results are highlighted in bold and italic , respectively.

\begin{table}[t]
\centering
\caption{Network settings of experimented baseline methods and our proposed framework.}
\resizebox{\columnwidth}{!}{
\begin{tabular}{lcc}\hline
Method	           &Spatial block layers &Temporal block layers \\ \hline
CNN+LSTM & 8 convolutional  & 3 LSTM \\
CNN+Bi-LSTM & 8 convolutional & 6 LSTM (3 forward and 3 backward)\\
CNN+GRU & 8 convolutional  & 3 GRU \\
CNN+Bi-GRU & 8 convolutional & 6 GRU (3 forward and 3 backward)\\
DA-CNN+Bi-GRU & 12 convolutional (8 convolutional and 4 attentional)   & 6 GRU (3 forward and 3 backward)\\ \hline
\end{tabular}
}
\label{tab:table2}
\end{table}

\begin{table}[t]
\centering
\caption{Quantitative comparative analysis of our proposed framework with other baseline methods.}
\resizebox{\columnwidth}{!}{
\begin{tabular}{lcc}\hline
Method	           &Dataset &Accuracy ($\%$)\\ \hline
CNN+LSTM  & YouTube action   &64.7\\
CNN+Bi-LSTM  & YouTube action   &84.2\\
CNN+GRU  & YouTube action   &88.5\\
CNN+Bi-GRU  & YouTube action   & \textit{92.1}\\
DA-CNN+Bi-GRU (Proposed)  & YouTube action   &\textbf{98.0}\\ \hline
CNN+LSTM  & UCF50   &76.3\\
CNN+Bi-LSTM  & UCF50  &83.3\\
CNN+GRU  & UCF50   &87.6\\
CNN+Bi-GRU  & UCF50   & \textit{93.6}\\
DA-CNN+Bi-GRU (Proposed) & UCF50   &\textbf{97.5}\\ \hline
CNN+LSTM  & HMDB51   &56.7\\
CNN+Bi-LSTM  & HMDB51  &63.2\\
CNN+GRU  & HMDB51   &68.0\\
CNN+Bi-GRU  & HMDB51   & \textit{72.4}\\
DA-CNN+Bi-GRU (Proposed)  & HMDB51   &\textbf{79.3}\\
\hline
\end{tabular}
}
\label{tab:table3}
\end{table}

\subsection{Comparison with State-of-the-Art Methods}
To show the effectiveness of our proposed framework for the human activity recognition task, we have conducted extensive comparative analysis of our method with the state-of-the-art methods in terms of overall accuracy. The quantitative comparisons of our method with the state-of-the-art methods for YouTube action, UCF50, and HMDB51 datasets are listed in Table \ref{tab:table4}, \ref{tab:table5}, and \ref{tab:table6}, respectively. The best results in these tables are represented in bold, whereas the runner-up results are highlighted in italic text. Considering the presented results, it can be noticed that our proposed framework (DA-CNN+Bi-GRU) outperforms state-of-the-art methods on UCF50 and HMDB51 datasets, whereas it attains runner-up performance on YouTube action dataset. For YouTube action dataset, the STDN \cite{zhang2020human} has the best performance with an accuracy of 98.2\%, whereas the proposed method attains the runner-up performance by obtaining an accuracy of 98.0\%, which is within 0.2\% accuracy of the best-performing STDN \cite{zhang2020human}. Thus, for most practical purposes, our proposed framework attains comparable performance to the STDN \cite{zhang2020human}. Rest of the methods that include multi-task hierarchical clustering \cite{liu2016hierarchical}, BT-LSTM \cite{ye2018learning}, deep autoencoder \cite{ullah2019action}, two-stream attention LSTM \cite{dai2020human}, weighted entropy-variance based feature selection \cite{afza2021framework}, dilated CNN+BiLSTM+RB \cite{muhammad2021human}, DS-GRU \cite{ullah2021efficient}, and local-global features + QSVM \cite{al2021making} obtain 89.7\%, 85.3\%, 96.2\%, 96.9\%, 94.5\%, 89.0\%, 97.1\%, and 82.6\% accuracies, respectively. For the UCF50 dataset, the proposed method dominates the state-of-the-art methods by obtaining the best accuracy of 97.5\%, whereas the (LD-BF) + (LD-DF) \cite{du2022linear} obtains the second-based accuracy of 96.7\%. The local-global features + QSVM \cite{al2021making} achieves the lowest accuracy of 69.4\%, whereas the rest of the methods including multi-task hierarchical clustering \cite{liu2016hierarchical}, deep autoencoder \cite{ullah2019action}, ensemble model with sward-based optimization \cite{zhang2021intelligent}, and DS-GRU \cite{ullah2021efficient} obtain 93.2\%, 96.4\%, 92.2\%, and 95.2\% accuracies, respectively. Finally, for the HMDB51 dataset comprising of challenging action videos, our proposed method achieves the best results by obtaining an accuracy of 79.3\%, whereas the runner-up method is evidential deep learning \cite{bao2021evidential} that attains an accuracy of 77.0\%. The multi-task hierarchical clustering method \cite{liu2016hierarchical} achieves an accuracy of 51.4\%, which is the lowest among all comparative methods on HMDB51 dataset. The rest of comparative methods including STPP+LSTM \cite{wang2017two}, optical flow + multi-layer LSTM \cite{ullah2018activity}, TSN \cite{wang2018temporal}, IP-LSTM \cite{yu2019learning}, deep autoencoder \cite{ullah2019action}, TS-LSTM + temporal-inception \cite{ma2019ts}, HATNet \cite{diba2019holistic}, correlational CNN+LSTM \cite{majd2020correlational}, STDN \cite{zhang2020human}, DB-LSTM+SSPF \cite{he2021db}, DS-GRU \cite{ullah2021efficient}, TCLC \cite{zhu2021temporal}, and semi-supervised temporal gradient learning \cite{xiao2022learning} obtain accuracies of 70.5\%, 72.2\%, 70.7\%, 58.6\%, 70.3\%, 69.0\%, 74.8\%, 66.2\%, 56.5\%, 75.1\%, 72.3\%, 71.5\%, and 75.9\%, respectively. Considering the overall comparative analysis, the proposed method obtains comparable performance to the best-performing method on the YouTube action dataset, and greatly dominates the state-of-the-art comparative methods on UCF50 and HMDB51 datasets, thus demonstrating the superiority of our proposed method over the exiting action recognition methods.

\begin{table}[t]
\centering
\caption{Quantitative comparative analysis of our proposed method with the state-of-the-art action recognition methods for YouTube action dataset.}
\resizebox{\columnwidth}{!}{
\begin{tabular}{lcc}\hline
Method	           &Year &Accuracy ($\%$)\\ \hline
Multi-task hierarchical clustering \cite{liu2016hierarchical} &2017   &89.7 \\
BT-LSTM \cite{ye2018learning} &2018   &85.3 \\
Deep autoencoder  \cite{ullah2019action} &2019   &96.2 \\
STDN   \cite{zhang2020human} &2020   &\textbf{98.2} \\
Two-stream attention LSTM \cite{dai2020human} &2020   &96.9 \\
Weighted entropy-variances based\\ feature selection \cite{afza2021framework} &2021   &94.5\\
Dilated CNN+BiLSTM+RB \cite{muhammad2021human} &2021  &89.0 \\
DS-GRU  \cite{ullah2021efficient} &2021  &97.1 \\
Local-global features + QSVM \cite{al2021making} &2021  &82.6 \\
DA-CNN+Bi-GRU (Proposed)  &2022  & \textit{98.0} \\\hline
\end{tabular}
}
\label{tab:table4}
\end{table}

\begin{table}[t]
\centering
\caption{Quantitative comparative analysis of our proposed method with the state-of-the-art action recognition methods for UCF50 dataset.}
\resizebox{\columnwidth}{!}{
\begin{tabular}{lcc}\hline
Method	           &Year &Accuracy ($\%$)\\ \hline
Multi-task hierarchical clustering\cite{liu2016hierarchical} &2017   &93.2 \\
Deep autoencoder  \cite{ullah2019action} &2019   &96.4 \\
Ensemble model with swarm-based\\ optimization  \cite{zhang2021intelligent} &2021   &92.2 \\
DS-GRU  \cite{ullah2021efficient} &2021  &95.2 \\
Local-global features + QSVM \cite{al2021making} &2021  &69.4 \\
(LD-BF) + (LD-DF)  \cite{du2022linear} &2022  & \textit{97.5} \\
DA-CNN+Bi-GRU (Proposed)  &2022  &\textbf{98.0} \\
\hline
\end{tabular}
}
\label{tab:table5}
\end{table}

\begin{table}[t]
\centering
\caption{Quantitative comparative analysis of our proposed method with the state-of-the-art action recognition methods for HMDB51 dataset.}
\resizebox{\columnwidth}{!}{
\begin{tabular}{lcc}\hline
Method	           &Year &Accuracy ($\%$)\\ \hline
Multi-task hierarchical clustering\cite{liu2016hierarchical} & 2017   & 51.4 \\
STPP+LSTM   \cite{wang2017two} & 2017   & 70.5 \\
Optical flow + multi-layer LSTM \cite{ullah2018activity} & 2018   & 72.2 \\
TSN \cite{wang2018temporal} & 2018  & 70.7 \\
IP-LSTM  \cite{yu2019learning} & 2019  & 58.6 \\
Deep autoencoder  \cite{ullah2019action} & 2019   & 70.3 \\
TS-LSTM + temporal-inception  \cite{ma2019ts} & 2019   & 69.0 \\
HATNet \cite{diba2019holistic} & 2019   & 74.8 \\
Correlational CNN + LSTM   \cite{majd2020correlational} & 2020   & 66.2 \\
STDAN  \cite{zhang2020human} & 2020   & 56.5 \\
DB-LSTM+SSPF \cite{he2021db} & 2021  & 75.1 \\
DS-GRU  \cite{ullah2021efficient} & 2021  &72.3 \\
TCLC   \cite{zhu2021temporal} & 2021  &71.5 \\
Evidential deep learning  \cite{bao2021evidential} & 2021  & \textit{77.0} \\
Semi-supervised temporal gradient\\ learning \cite{xiao2022learning} & 2022  & 75.9 \\
DA-CNN+Bi-GRU (Proposed)  & 2022  &\textbf{79.3} \\
\hline
\end{tabular}
}
\label{tab:table6}
\end{table}

\begin{figure*}[t]
    \includegraphics[width=181mm]{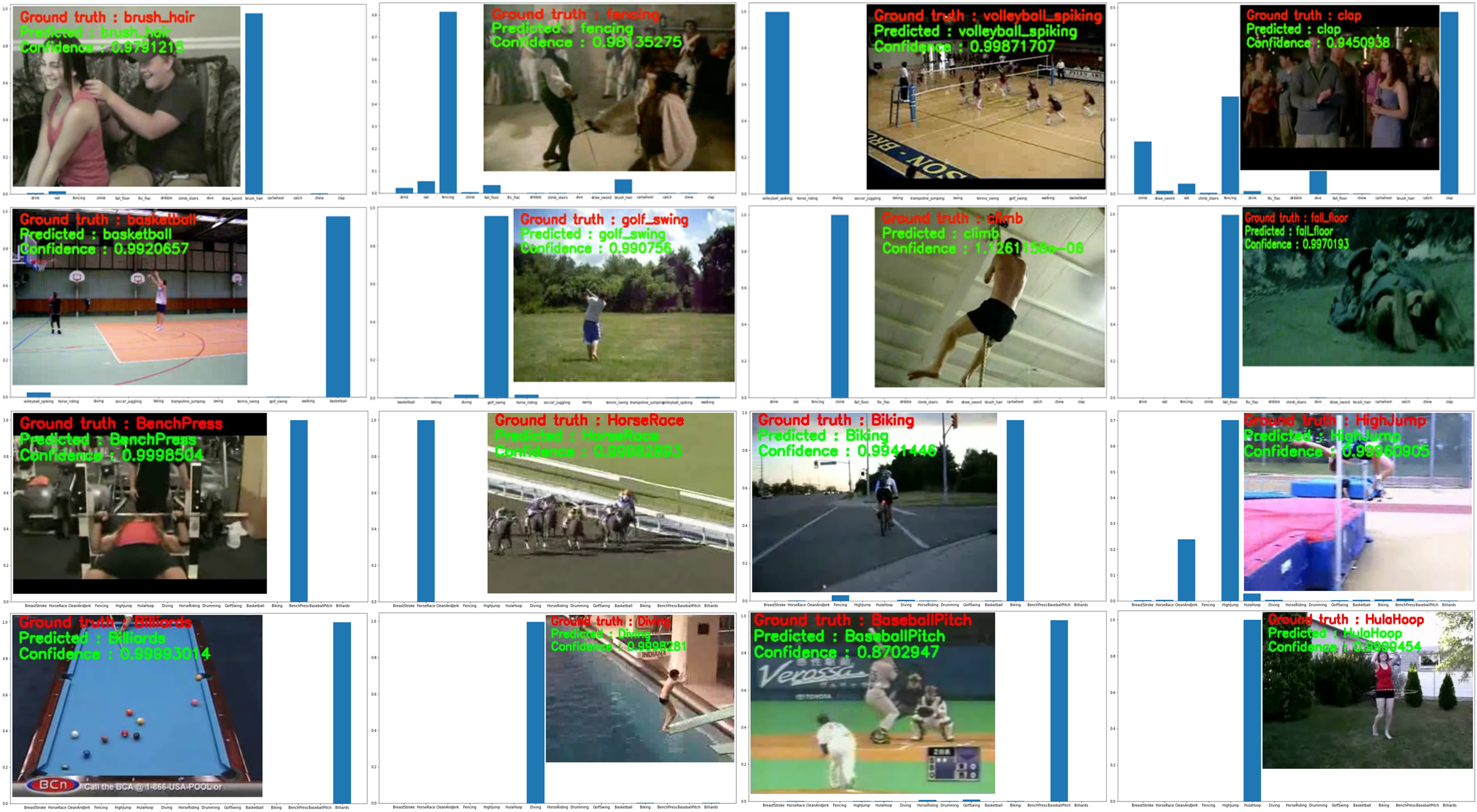}
    \caption{The visual recognition results of our proposed DA-CNN+Bi-GRU framework with predicted classes and their confidence scores for the test videos taken from the YouTube action, UCF50, and HMDB51 datasets.}
    \label{Fig:VisualRecognition}
\end{figure*}

\subsection{Action Recognition Visualization}
To validate the recognition efficiency of our proposed framework, we have tested our framework on 15\% of test videos taken from each dataset (including YouTube action, UCF50, and HMDB51). The prepared test sets are validated for the action recognition task using our proposed framework and the visual results from the test experiments are depicted in Figure~\ref{Fig:VisualRecognition}. In Figure~\ref{Fig:VisualRecognition}, the representative frames of the  predicted action clips are presented along with their ground truths, model predicted actions, and confidence scores over the probability prediction bar graphs for better understanding of readers. It can be perceived from the presented visual results that the proposed framework predicts most of the actions including brush hair, volleyball spiking, basketball, climb, fall floor, bench press, horse race, billiards, diving, baseball pitch, and hula hoop with 0.99\sout{\%} probability or 99\% confidence. Though, for some action classes, such as clap, fencing, golf swing, and high jump, the proposed framework also generates non-zero probabilities for wrong action classes; however, these probabilities for wrong action classes are still very low and thus do not affect the prediction of actual action class. Hence, the obtained qualitative visual results verify the effectiveness of our proposed framework for practical use in different vision-based human action recognition and monitoring environments.

\subsection{Runtime Analysis}
To analyze the effectiveness and feasibility of our proposed framework for practical applications in real-time environments, we have estimated the runtime of our method for action recognition tasks in terms of SPF and FPS with and without using GPU resources. The obtained runtime results are then compared with the stat-of-the-art methods. Table~\ref{tab:Runtime} presents and compares the runtime of our proposed framework with the running times of the contemporary action recognition methods. Results in Table~\ref{tab:Runtime} demonstrate that our proposed framework outperforms the state-of-the-art methods when executing on both GPU and central processing unit (CPU) platforms. Results indicate that our proposed framework attains 0.0036 SPF and 300 FPS while running on GPU, whereas it attains 0.0049 SPF and 250 FPS while running on CPU. Results further show that the second-best execution time results on GPU are achieved by \cite{wang2017two}, which are 0.0053 SPF and 186.6 FPS. In Table~\ref{tab:Runtime}, the best runtime results are highlighted with bold and runner-up results are emphasized with italic. Experimental results indicate that for the SPF metric, our proposed framework can provide an improvement of up to 18.6$\times$ when running on GPU and an improvement of 87.76$\times$ when running on CPU as compared to other contemporary activity recognition methods. Experimental results further reveal that for the FPS metric, our proposed framework can provide an improvement of up to 21.43$\times$ when running on GPU and an improvement of 166.6$\times$ when running on CPU as compared to other contemporary activity recognition methods. It is also worth mentioning here that the storage requirement of our proposed framework is just 5.4\,MB, and thus our framework can be run on resource-constrained IoT and edge devices with very limited memory including today’s smart cameras, Arduino, and Raspberry pi. These runtime and storage requirement results demonstrate that the proposed framework is a suitable candidate for deployment on resource-constrained IoT and edge devices as the proposed framework exhibits better accuracy, lower execution time, and low storage requirements as compared to contemporary activity recognition methods.

\begin{table}[t!] 
%\rowcolors{3}{gray!25}{white}
\caption{Runtime analysis of our proposed framework with state-of-the-art human action recognition methods.}
\resizebox{\columnwidth}{!}{
\begin{tabular}{cccccccc}
\toprule
\multirow{2}{*}{Method} & \multicolumn{2}{c}{Seconds per Frame (SPF)} & \multirow{2}{*}{Year} & \multicolumn{2}{c}{Frames per Second (FPS)} & \\
\cmidrule{2-3} \cmidrule{5-6}
                        & GPU & CPU &                       & GPU        & CPU        \\ \midrule
STPP+LSTM \cite{wang2017two}  & \textit{0.0053} & - & 2017 & \textit{186.6} & - \\  
Optical flow + multi-layer LSTM \cite{ullah2018activity}  &0.0356 & 0.18 & 2018 & 30 & 3.5 \\ 
Deep autoencoder \cite{ullah2019action}  & 0.0430 & 0.43 & 2019 & 24 & 1.5 \\ 
IP-LSTM \cite{yu2019learning}  & 0.0431 & - & 2019 & 23.2 & - \\ 
STDN  \cite{zhang2020human}  & 0.0075 & - & 2020 & 132 & - \\ 
DS-GRU \cite{ullah2021efficient}  & 0.0276 & - & 2021 & 37 & - \\ 
(LD-BF) + (LD-DF) \cite{du2022linear}  & 0.0670 & - & 2022 & 14 & - \\
DA-CNN+Bi-GRU (Proposed) & \textbf{0.0036} & \textbf{0.0049} & 2022 & \textbf{300} & \textbf{250} \\
            
            \\ \bottomrule
\end{tabular}
}
\label{tab:Runtime}
\end{table}

\section{Conclusions and Future Research Directions} \label{sec:conclusion}
In this work, we have proposed a cascaded spatial-temporal discriminative feature learning framework for human activity recognition in video streams. The proposed method encapsulates the attentional (channel and spatial attention) CNN architecture and bi-directional GRU network as a unified framework for single instance training and efficient spatial temporal modeling of human actions. The attentional CNN architecture comprises of channel and spatial attentions, which help retrieve the prominent discriminative features from the object-specific regions, and thus generate high quality saliency-aware feature maps. The bi-directional GRU learns the temporal modeling of long-term human action sequences using two-way gradient learning (i.e., forward and backward pass), which allows our approach to utilize the learned knowledge not only from the previous frames but also from the upcoming/next frames. Such bi-directional modeling of human actions greatly helps our method to improve the learning ability while training and the prediction precision while inferencing. To evaluate the efficiency of our method, we have conducted extensive experiments on three publicly available human action benchmark datasets. The obtained experimental results are compared with the state-of-the-art methods on three benchmark human action recognition datasets, that include youtube action, UCF50, and HMDB51 datasets. Experimental results verify the effectiveness of our method in terms of both model robustness and computational efficiency. Further, we have analyzed the runtime performance of our proposed framework in terms of seconds per frame (SPF) and frames per second (FPS) for both CPU and GPU execution environments. The obtained runtime assessment results reveal that our proposed framework can attain an improvement of up to 88$\times$ for the SPF metric and up to 167$\times$ for the FPS metric as compared to other contemporary action recognition methods. Additionally, our proposed framework requires a storage of only 5.3\,MB, which makes it feasible for deployment on devices with limited memory. Thus, the overall efficiency of our framework in terms of recognition performance (accuracy), low execution time, and low storage requirements, makes our framework a strong candidate for real-time IoT and edge applications.

Currently, our proposed method only uses spatial attention (channel and spatial attention) mechanism, which is indeed very effective. However, in future we plan to use temporal attention mechanism together with spatial attention, because such hybrid attention has a great potential to improve the human activity recognition performance.      

\bibliographystyle{IEEEtran}
\balance
\bibliography{References}

% that's all folks

\end{document}